\documentclass{article}

\usepackage[preprint]{neurips_2026}

\pdfoutput=1

\usepackage[utf8]{inputenc}
\usepackage[T1]{fontenc}

\usepackage{hyperref}
\usepackage{url}
\usepackage{booktabs}
\usepackage{amsfonts}
\usepackage{nicefrac}
\usepackage{microtype}
\usepackage{graphicx}
\usepackage{amsmath}
\usepackage{algorithm}
\usepackage{algpseudocode}

\usepackage{xcolor}
\title{NeuroPlastic: A Plasticity-Modulated Optimizer for\\
Biologically Inspired Learning Dynamics}

\begin{document}

\author{
\textbf{Douglas Jiang$^{1,2,5}$, Yuechen Wang$^{1,5}$, 
Jiayi Wang$^{1,3}$, Jiaying Geng$^{1,4}$,}\\
\textbf{Qinglong Wang$^{1}$, and Feng Tian$^{1}$\textsuperscript{*}}\\
[0.9em]
$^{1}$Department of Neurology, Beth Israel Deaconess Medical Center, Harvard Medical School,\\
Boston, MA 02115, USA\\
$^{2}$Department of Biostatistics, Harvard T.H. Chan School of Public Health,\\
Boston, MA 02115, USA\\
$^{3}$Department of Statistics, John A. Paulson School of Engineering and Applied Sciences,\\
Harvard University, Cambridge, MA 02138, USA\\
$^{4}$Department of Psychology, Columbia University, New York, NY 10027, USA\\
$^{5}$Equal contributions\\[0.5em]
$^{*}$Correspondence: \texttt{ftian@bidmc.harvard.edu}
}

\maketitle

\begin{abstract}
Optimization algorithms are fundamental to modern deep learning, yet most widely used methods rely on update rules based primarily on local gradient statistics. We introduce \textbf{NeuroPlastic}, a plasticity-modulated optimizer that augments gradient-based updates with an adaptive multi-signal modulation mechanism inspired by multi-factor synaptic plasticity, a concept from neurobiology. NeuroPlastic dynamically scales gradient updates using interacting components that capture gradient, activity-like, and memory-like statistics, forming a lightweight modulation layer compatible with standard deep learning training pipelines. Across image classification benchmarks, NeuroPlastic consistently improves over a controlled gradient-only ablation, with more pronounced gains on the Fashion-MNIST benchmark and in reduced-data regimes. In transfer experiments on CIFAR-10 with ResNet-18, the method remains stable and competitive without retuning. These results suggest that multi-signal plasticity-inspired modulation can provide a useful extension to conventional gradient-driven optimization, particularly when learning signals are limited or noisy, and offer a promising direction for gradient-based methods in deep learning. Code is available at: https://github.com/Sunnylincc/NeuroPlastic-Optimizer.
\end{abstract}

\section{Introduction}
\noindent
Optimization algorithms are a foundation of modern deep learning. The ability of neural networks to generalize depends not only on model architecture and training data, but critically on how parameters are updated during training~\citep{goodfellow2016deep}. Over the past decade, stochastic gradient descent and adaptive methods such as Adam and AdamW have become the dominant tools for this purpose~\citep{robbins1951stochastic,kingma2015adam,loshchilov2019adamw}. They demonstrate strong empirical performance across vision, language, and large-scale foundation model training \citep{he2016resnet,brown2020,vaswani2017attention}. These methods share a common design principle: parameter updates are computed as functions of local gradient statistics, either raw stochastic gradients with momentum, or first- and second-order exponential moving averages~\citep{polyak1964momentum,rumelhart1986learning,kingma2015adam}. This design is computationally efficient and broadly effective, but it also constrains the learning signal. In all cases, the effective learning signal at each step is derived from gradient information as a single source, with history encoded only through moving averages of that same signal. When gradients are informative and plentiful, this is sufficient. When learning signals are weak, sparse, or noisy, single-signal updates may limit the model's ability to adapt effectively~\citep{fremaux2016threefactor}.

\par\vspace{1em}
Unlike stochastic gradients, biological neural systems like the human brain organize learning through multiple interacting components. Classical Hebbian learning describes synaptic strengthening as a function of correlated pre- and post-synaptic activity~\citep{hebb1949organization}. Subsequent work in computational neuroscience extended this idea into three-factor or multi-factor plasticity rules, where synaptic updates depend not only on co-activation but also on additional signals~\citep{gerstner2011hebbian,lillicrap2020backprop}. These signals may include neuromodulators, activity traces reflecting the history of a neuron's firing, and network-level regulatory mechanisms that maintain stability over time~\citep{hebb1949organization,fremaux2016threefactor}. Together, these factors allow learning to be modulated by context: what has recently occurred, what the network is currently doing, and signals that reflect the broader relevance of a given update. The key feature of these rules is that the effective learning signal at any synapse emerges from multiple interacting sources rather than a single scalar. This biological architecture suggests a design principle: learning signals constructed from multiple interacting components may be more informative than gradient information alone, particularly in regimes where the primary learning signal is degraded.

\par\vspace{1em}
However, translating these neurobiological principles into scalable deep learning optimizers remains challenging, as neural plasticity mechanisms are not readily compatible with gradient-based training in modern deep learning architectures. Therefore, we ask a focused question: can multi-factor plasticity be implemented in a lightweight optimizer for standard deep learning settings?

\par\vspace{1em}
To investigate this question, we introduce \textbf{NeuroPlastic}. NeuroPlastic augments gradient updates with a plasticity modulation coefficient constructed from three normalized signals: gradient magnitude, an exponential moving average of gradient activity, and an Adam-style memory term, specifically the ratio of first- to second-moment estimates followed by population-level normalization. These three signals jointly determine a modulation factor that scales the effective gradient update. We evaluate the NeuroPlastic optimizer against gradient-only and standard optimizers (SGD, Adam, AdamW) on image classification benchmarks of increasing difficulty. Results show consistent improvements over the gradient-only ablation on MNIST~\citep{lecun1998gradient}, with clearer gains on Fashion-MNIST~\citep{xiao2017fashion}. On CIFAR-10 with ResNet-18~\citep{krizhevsky2009learning,he2016resnet}, the method remains stable and competitive without retuning.

\par\vspace{1em}
Our main contributions can be summarized as follows:

\begin{enumerate}
    \item We introduce \textbf{NeuroPlastic}, a biologically inspired optimizer that modulates gradient updates through a plasticity signal combining gradient information with activity- and memory-like components.

    \item We design a stabilization mechanism that regulates update magnitudes and maintains stable optimization dynamics across learning rates.

    \item We develop a reproducible benchmarking framework and diagnostic tools for analyzing plasticity-modulated optimization behavior during training.

    \item We empirically evaluate NeuroPlastic across standard benchmarks and data regimes, demonstrating consistent improvements over gradient-only updates, with the advantage scaling with task difficulty and data scarcity.
    
\end{enumerate}

\section{Related Work}
Despite decades of progress in optimizer design, the question of what constitutes a sufficient learning signal remains largely unexamined. Existing optimizers treat gradient information as the only basis for parameter updates. While richer update rules have been explored through meta-learning, second-order methods, and biologically inspired plasticity, none has produced a practical, drop-in alternative competitive with Adam that incorporates multi-signal modulation. We review these three lines of work below, identifying their limitations and how NeuroPlastic extends them to address these gaps.

\subsection{Gradient-Based Optimization in Deep Learning}

Stochastic gradient descent and its momentum-based extensions remain the most widely used optimization algorithms for neural network training. Adam generalizes this by computing adaptive per-parameter learning rates using exponential moving averages of first and second gradient moments, and AdamW decouples weight decay to improve generalization~\citep{kingma2015adam,loshchilov2019adamw}. 
Other adaptive methods including AdaGrad, RMSProp, and AdaDelta similarly rely on accumulated gradient statistics to scale updates~\citep{duchi2011adagrad,zeiler2012adadelta,tieleman2012rmsprop}. These methods are computationally efficient and have proven to be effective in many large-scale machine learning applications, including computer vision and natural language processing~\citep{kingma2015adam,loshchilov2019adamw,duchi2011adagrad,tieleman2012rmsprop}. However, they share a structural limitation relevant to our work: the adaptive component of the update rule is derived entirely from transformations of the gradient signal. History enters only through moving averages of gradient quantities, and the effective learning signal at each step is determined by gradient information alone. NeuroPlastic is designed to explore the possibility of constructing a modulation signal from multiple sources.

\subsection{Alternative and Learned Optimizers}

Beyond classical adaptive gradient methods, a variety of alternative optimization strategies have been proposed to enrich or replace gradient-based update rules. Second-order and preconditioned methods such as K-FAC and Shampoo incorporate curvature information to accelerate convergence, often at the cost of significant computational overhead~\citep{eschenhagen2023kfac,martens2015curvature,martens2010hessianfree,gupta2018shampoo}. A separate line of work attempts to learn update rules directly from data: meta-learning approaches train optimizers to generalize across tasks, but typically require expensive meta-training and do not transfer reliably outside the training distribution~\citep{andrychowicz2016,chandra2022}. More recently, symbolic discovery approaches such as Lion have searched mathematical expressions to identify compact and effective update rules, resulting in lightweight alternatives to Adam~\citep{chen2023symbolic,wang2022enos,berthet2020,joseph2023localsgd}. 
Together, these directions demonstrate that the space of useful optimizer designs extends well beyond first-order gradient descent and that departing from standard update rules in principled ways can yield practical benefits. NeuroPlastic is motivated by the same observation but takes a different path. Rather than learning or searching for an update rule, we derive one from the functional structure of multi-factor biological plasticity. The result is a modulation mechanism that is fixed in structure but adaptive in its per-step behavior.

\subsection{Biologically Inspired Learning Rules}

The computational neuroscience literature has developed a rich family of learning rules motivated by synaptic plasticity mechanisms in biological neural circuits. Hebbian learning, the first formalization, shows that synaptic weights increase in proportion to the correlated activity of pre- and post-synaptic neurons~\citep{hebb1949organization}. 
Three-factor plasticity rules extend this framework by introducing a third modulatory signal that influences synaptic updates~\citep{gerstner2018threefactor}. 
These additional signals may include a neuromodulatory input that encodes reward, novelty, or task relevance, determining whether a Hebbian update is applied and at what magnitude~\citep{hebb1949organization,gerstner2011hebbian,fremaux2016threefactor}. The result is an update rule in which the effective learning signal is a product of three interacting terms: a local activity-dependent term, a synaptic activity trace encoding the history of co-activation, and a global or semi-global modulatory signal.

Several lines of work have explored whether biologically motivated plasticity rules can be incorporated into machine learning systems~\citep{lillicrap2020backprop,marblestone2016,richards2019dendritic}. The most prominent efforts focused on making credit assignment more biologically plausible. Feedback alignment and its direct variant replace the standard backward pass of backpropagation with fixed random feedback connections~\citep{lillicrap2016feedback, nokland2016direct}. While this produces a more biologically grounded update rule, performance degrades on tasks beyond simple benchmarks, and the approach has not been shown to scale to modern architectures. Target propagation methods, such as difference target propagation~\citep{lee2015difference}, take a different route by computing layer-wise learning targets rather than gradients, a design with natural connections to local plasticity rules. However, they often suffer from inefficient parameter updates in typical deep networks and similarly fall short of standard backpropagation performance at scale~\citep{meulemans2020theoretical}.

A closer parallel to our work is differentiable plasticity: prior work has shown that Hebbian plastic connections, where each synapse carries both a stable weight and an activity-dependent component, can be optimized end-to-end by gradient descent~\citep{miconi2018differentiable}. Subsequent work extended this to neuromodulated plasticity, where a learned modulatory signal gates the Hebbian update moment-to-moment, yielding improvements on meta-learning and reinforcement learning tasks~\citep{miconi2019backpropamine}. These approaches share the core intuition motivating NeuroPlastic, but they are primarily evaluated on small recurrent architectures in meta-learning or continual learning settings, and are not designed as drop-in replacements for standard optimizers. No prior work has instantiated multi-signal modulation directly as a parameter update rule and evaluated it against a controlled gradient-only ablation in standard deep learning settings.

\section{Methods}

\subsection{Overview}

We propose \textbf{NeuroPlastic}, a gradient-based optimizer in which the raw gradient is modulated by a plasticity coefficient constructed from multiple normalized optimization signals and then stabilized before being applied. Let $\theta_t$ denote the model parameters at step $t$, and let
\begin{equation}
g_t = \nabla_{\theta}\mathcal{L}(\theta_t)
\end{equation}
denote the stochastic gradient. The core update takes the form
\begin{equation}
\theta_{t+1} = \theta_t - \eta_t \, \widetilde{u}_t,
\end{equation}
where $\eta_t$ is the effective learning rate and $\widetilde{u}_t$ is a stabilized, plasticity-modulated update.

In the default rule-based configuration, NeuroPlastic first computes a per-parameter modulation factor $\alpha_t$, forms the modulated update

\begin{equation}
u_t = \alpha_t \odot g_t
\end{equation}
and then applies a stabilization operator
\begin{equation}
\widetilde{u}_t = \mathcal{S}(u_t).
\end{equation}

\begin{figure}
\centering
\includegraphics[width=\linewidth]{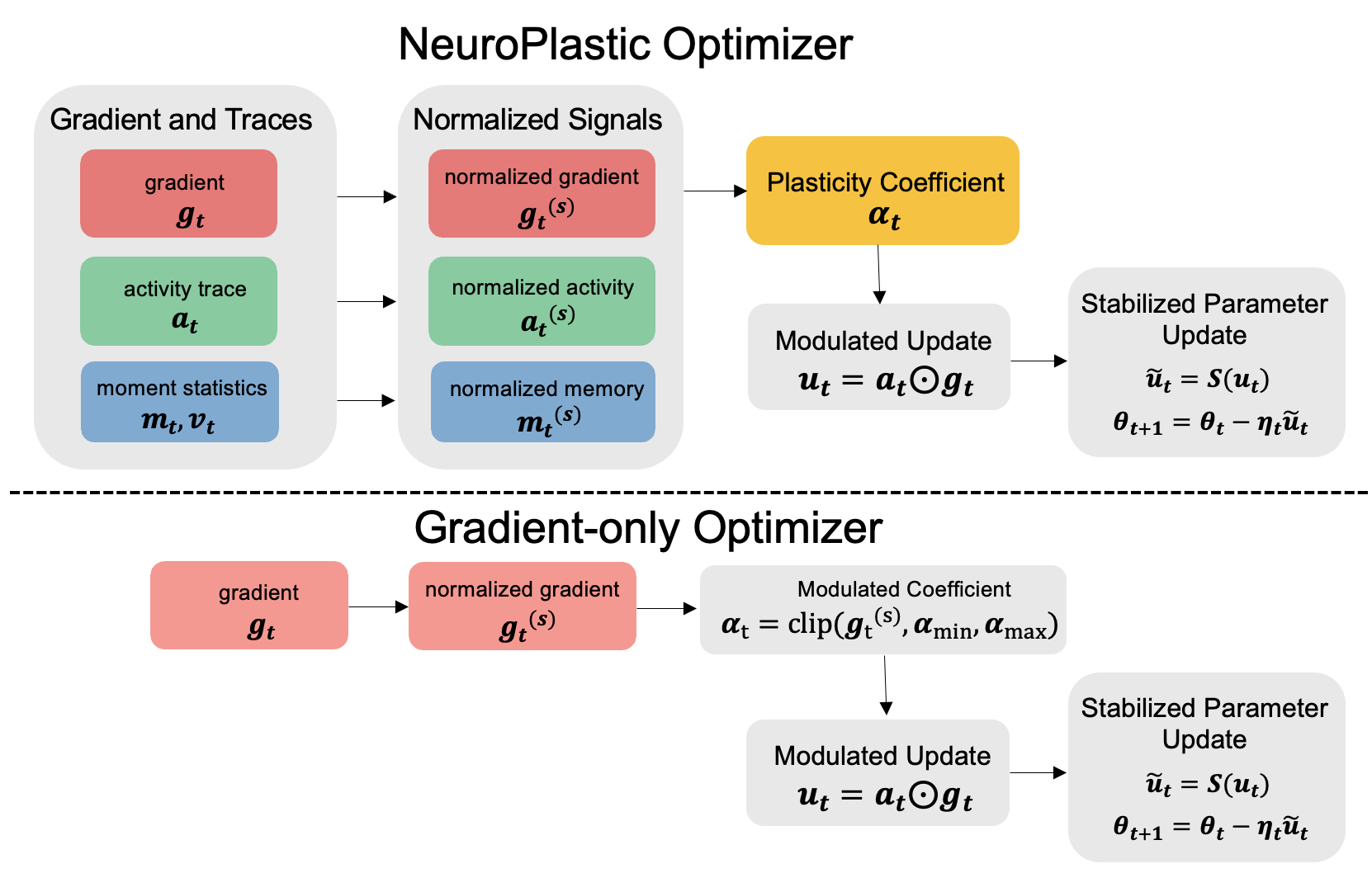}

\caption{
\textbf{Overview of the NeuroPlastic optimizer.} Gradient updates are modulated by a plasticity coefficient constructed from normalized gradient, activity-like, and memory-like signals. The resulting modulated update is
$u_t = \alpha_t \odot g_t$, which is then stabilized to produce $\widetilde{u}_t = \mathcal{S}(u_t)$ before the parameter update $\theta_{t+1} = \theta_t - \eta_t \widetilde{u}_t$. The gradient-only ablation removes the activity and memory contributions while retaining gradient-based modulation, using $\alpha_t^{\mathrm{grad}} = \mathrm{clip}(g_t^{(s)}, \alpha_{\min}, \alpha_{\max})$.
}
\label{fig:neuroplastic_architecture}
\end{figure}
\subsection{Activity and Memory Signals}

NeuroPlastic maintains three internal statistics derived from the gradient: an activity trace, a first-moment trace, and a second-moment trace.

\paragraph{Activity trace.}
The activity trace is an exponential moving average of gradient magnitudes:
\begin{equation}
a_t = \beta_a a_{t-1} + (1-\beta_a)|g_t|.
\end{equation}

\paragraph{Momentum and variance traces.}
We also maintain Adam-style first- and second-moment statistics:
\begin{align}
m_t &= \beta_1 m_{t-1} + (1-\beta_1)g_t, \\
v_t &= \beta_2 v_{t-1} + (1-\beta_2)(g_t \odot g_t).
\end{align}

\subsection{Normalized Plasticity Signals}

The optimizer constructs three normalized signals from the quantities above.

\paragraph{Gradient signal.}
\begin{equation}
g_t^{(s)} = \frac{|g_t|}{\mathrm{mean}(|g_t|)+\epsilon}.
\end{equation}

\paragraph{Activity signal.}
\begin{equation}
a_t^{(s)} = \frac{a_t}{\mathrm{mean}(a_t)+\epsilon}.
\end{equation}

\paragraph{Memory signal.}
The memory signal combines the first- and second-moment traces:
\begin{equation}
m_t^{(s)} =
\frac{|m_t|/(\sqrt{v_t}+\epsilon)}
{\mathrm{mean}\!\left(|m_t|/(\sqrt{v_t}+\epsilon)\right)+\epsilon},
\end{equation}
where $m_t$ and $v_t$ are the raw (non-bias-corrected) moment estimates 
as defined in Equations~(6)--(7), and $\epsilon > 0$ is a small constant 
for numerical stability.

\subsection{Plasticity Coefficient}

\paragraph{Gradient-only ablation.}
A key implementation detail is that the gradient-only ablation used in the code does \emph{not} set $\alpha_t=1$. Instead, it removes the activity and memory contributions while retaining gradient-based modulation:
\begin{equation}
\alpha_t^{\mathrm{grad}} = \mathrm{clip}\!\left(g_t^{(s)}, \alpha_{\min}, \alpha_{\max}\right).
\end{equation}

\paragraph{Full plasticity rule.}
In the full rule-based configuration, the plasticity coefficient is computed as a weighted combination of the three normalized signals:
\begin{equation}
\alpha_t^{\mathrm{full}}
=
w_a a_t^{(s)}
+
w_g g_t^{(s)}
+
w_m m_t^{(s)},
\end{equation}
followed by clipping:
\begin{equation}
\alpha_t^{\mathrm{full}}
\leftarrow
\mathrm{clip}\!\left(\alpha_t^{\mathrm{full}}, \alpha_{\min}, \alpha_{\max}\right).
\end{equation}

\paragraph{Warmup interpolation.}
The final plasticity coefficient interpolates between the gradient-only and full rules:
\begin{equation}
\alpha_t
=
\alpha_t^{\mathrm{grad}}
+
\lambda_t\!\left(\alpha_t^{\mathrm{full}}-\alpha_t^{\mathrm{grad}}\right),
\end{equation}
where
\begin{equation}
\lambda_t = s_{\mathrm{plast}} \cdot \gamma_t^{\mathrm{warmup}}.
\end{equation}
Here $s_{\mathrm{plast}}$ is a plasticity scaling factor and $\gamma_t^{\mathrm{warmup}}\in[0,1]$ is a warmup gate.

\paragraph{Parameter-wise versus scalar modulation.}
By default, $\alpha_t$ is computed per parameter. If parameter-wise modulation is disabled, the implementation replaces $\alpha_t$ by its mean value and broadcasts the resulting scalar across all coordinates of the tensor.

\subsection{Homeostatic Stabilization}

After constructing the plasticity-modulated update
\begin{equation}
u_t = \alpha_t \odot g_t,
\end{equation}
NeuroPlastic applies a homeostatic stabilization operator $\mathcal{S}(\cdot)$ consisting of two steps.

\paragraph{Step 1: Norm clipping.}
Let $\tau$ denote the maximum allowed update norm. If $\|u_t\|>\tau$, the update is rescaled:
\begin{equation}
u_t \leftarrow
u_t \cdot \frac{\tau}{\|u_t\|+\epsilon}.
\end{equation}

\paragraph{Step 2: RMS-based gain control.}
The update is then adjusted using its root-mean-square (RMS) magnitude:
\begin{equation}
\mathrm{RMS}(u_t)=\sqrt{\mathrm{mean}(u_t^2)+\epsilon}.
\end{equation}
Given a target RMS value $r_{\mathrm{target}}$, the stabilizer computes a gain
\begin{equation}
\gamma_t^{\mathrm{stab}}
=
1 + \rho\left(r_{\mathrm{target}}-\mathrm{RMS}(u_t)\right),
\end{equation}
and clips it to a bounded interval:
\begin{equation}
\gamma_t^{\mathrm{stab}}
\leftarrow
\mathrm{clip}\!\left(\gamma_t^{\mathrm{stab}}, \gamma_{\min}, \gamma_{\max}\right).
\end{equation}
The stabilized update is therefore
\begin{equation}
\widetilde{u}_t = \gamma_t^{\mathrm{stab}} u_t.
\end{equation}

The implementation also supports layer-dependent scaling of the target RMS through a multiplicative factor; when all scaling factors are set identically, the same target RMS is applied across parameter groups.

\subsection{Weight Decay and Effective Learning Rate}

The implementation supports both coupled and decoupled weight decay. In the coupled setting, the decay term is added directly to the update. In the decoupled setting, parameters are separately shrunk before the stabilized update is applied, analogous to AdamW-style regularization.

The optimizer also supports an optional learning-rate controller that rescales the base learning rate according to a controller signal derived from the discrepancy between $\alpha_t$ and $\alpha_t^{\mathrm{grad}}$. In the default configuration, this controller is disabled, so the effective learning rate reduces to the base learning rate:
\begin{equation}
\eta_t = \eta.
\end{equation}

\section{Experiments}

We evaluate NeuroPlastic through a series of controlled experiments designed not only to assess its empirical performance, but also to understand how plasticity provides useful inductive bias beyond standard gradient-based updates.

\subsection{MNIST and Fashion-MNIST}

We evaluate the optimizer on two standard image-classification benchmarks, using MNIST as a lightweight sanity check and Fashion-MNIST as a more challenging follow-up task. This paired analysis allows us to assess whether the contribution of plasticity-modulated updates changes as task difficulty increases.

\begin{figure}[t]
\centering
\includegraphics[width=\linewidth]{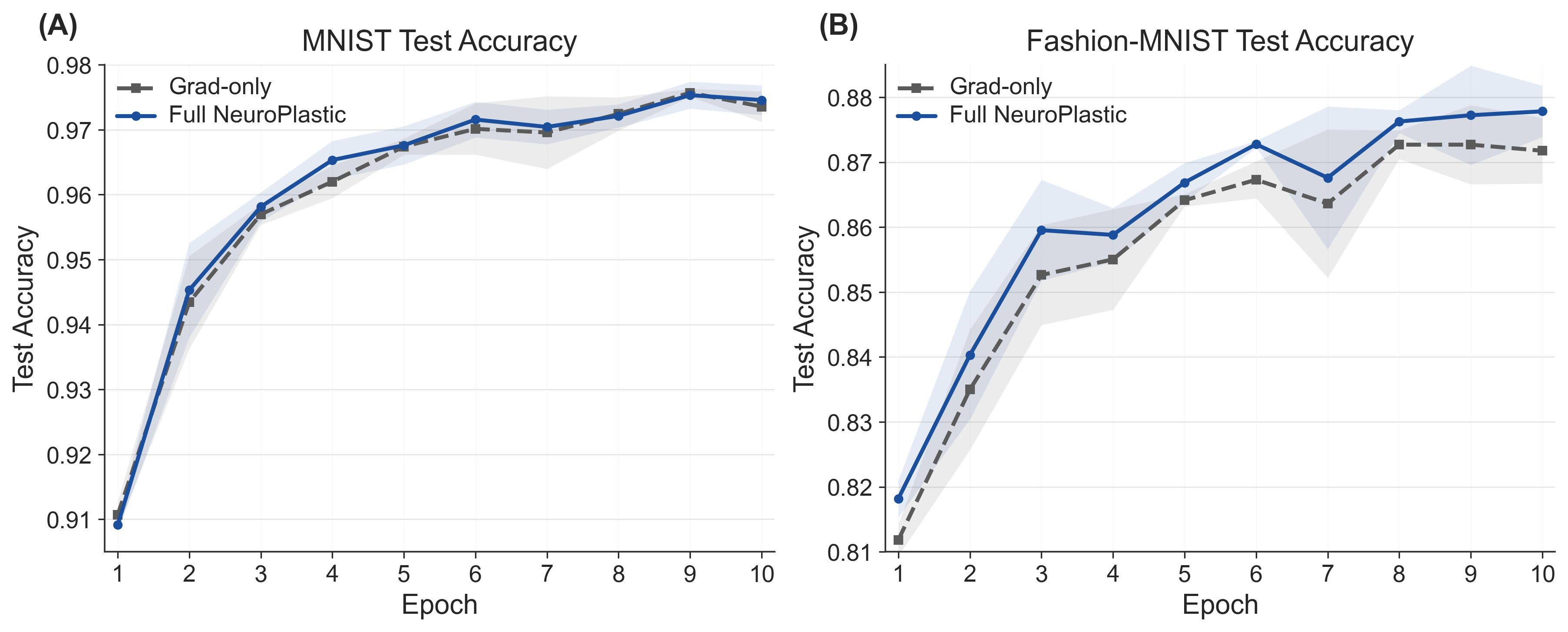}
\caption{
\textbf{Test accuracy across training epochs on MNIST and Fashion-MNIST.}
(A) MNIST test accuracy comparing the full NeuroPlastic optimizer with the gradient-only ablation.
(B) Fashion-MNIST test accuracy under the same training configuration.
Shaded regions denote the standard deviation across independent runs.
}
\label{fig:mnist_training}
\end{figure}

On MNIST, NeuroPlastic achieves slightly higher final accuracy and lower training loss than the gradient-only ablation across three training seeds. The improvement in final test accuracy is small (approximately +0.001 absolute), but is consistent across all seeds and comparisons. These results indicate that the plasticity-modulated update rule provides a modest but reproducible improvement over a purely gradient-driven update, thus offering a stable baseline for the more challenging evaluations that follow. Training curves further show stable optimization behavior throughout training (Figure~\ref{fig:mnist_training}A).

On Fashion-MNIST, NeuroPlastic demonstrates its advantages by consistently outperforming the gradient-only baseline on multiple metrics. Across three seeds, NeuroPlastic improves mean final test accuracy by approximately +0.006 compared with the gradient-only ablation and consistently achieves lower training loss. It outperforms the gradient-only baseline on all shared seeds for both final accuracy and best observed accuracy during training. These results suggest that the plasticity signal becomes more useful as task difficulty increases, while maintaining stable training dynamics (Figure~\ref{fig:mnist_training}B).

To understand why plasticity modulation improves performance on Fashion-MNIST, we examine the internal optimizer dynamics during training. As detailed in Appendix~\ref{appendix:mechanistic_analysis}, the full NeuroPlastic optimizer maintains a consistently higher and more stable plasticity coefficient throughout training, and produces a distinct effective update norm trajectory, with slightly larger early updates and a smoother late-stage decay relative to the gradient-only ablation. Critically, the raw gradient norm remains similar between the two conditions, suggesting that the gains arise from how updates are modulated and stabilized rather than from any change in gradient magnitude.

\subsection{Low-Data Regime Analysis}

To further examine when plasticity modulation is most beneficial, we conduct a controlled low-data study on Fashion-MNIST by training models using only a fraction of the available training data. We evaluate fractions of $\{0.1, 0.25, 0.5, 1.0\}$ of the original training set while keeping the test set unchanged. Across all data fractions, NeuroPlastic improves both final accuracy and training loss relative to the gradient-only ablation, with absolute gains in final accuracy ranging from approximately +0.004 to +0.006. Interestingly, the advantage of NeuroPlastic tends to slightly increase when less data is available, although the relationship is not strictly monotonic. These findings suggest that plasticity-modulated updates may be particularly beneficial when learning signals are weaker or data is limited (Figure~\ref{fig:lowdata_fashion}). A fuller characterization of how plasticity coefficients evolve across training is provided in Appendix~\ref{appendix:mechanistic_analysis}.

\begin{figure}[t]
\centering
\includegraphics[width=\linewidth]{figure/fig3_combined.png}
\caption{
\textbf{Low-data regime analysis on Fashion-MNIST.}
(A) Final test accuracy under different fractions of the training dataset.
Across all data fractions, the full NeuroPlastic optimizer outperforms the gradient-only ablation, with larger gains in the more data-constrained regimes.
(B) Final test loss under the same settings.
Plasticity-modulated updates consistently achieve lower loss across data fractions.
Shaded regions denote the standard deviation across independent runs.
}
\label{fig:lowdata_fashion}
\end{figure}

\subsection{Comparison with Standard Optimizer Baselines}
To contextualize NeuroPlastic under standard training settings, we include a comparison on CIFAR-10~\citep{krizhevsky2009learning} using ResNet-18 and standard optimizer baselines (SGD, Adam, AdamW). Because this experiment intentionally uses the MNIST-tuned configuration without retuning, the results should be interpreted primarily as a transferability test rather than a fully optimized comparison.

\begin{figure}[t]
    \centering
    \includegraphics[width=\linewidth]{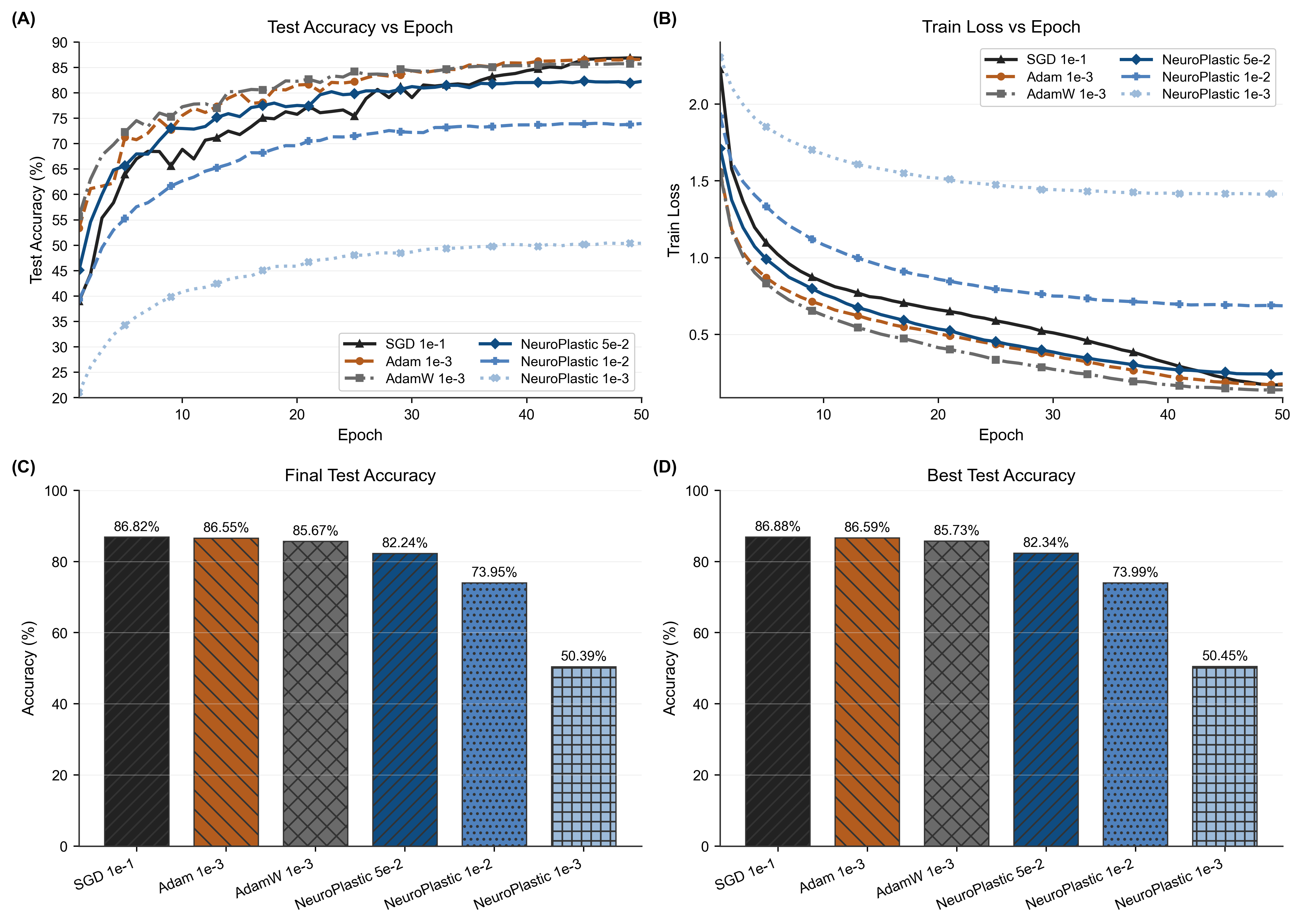}
    \caption{
    \textbf{Validation on CIFAR-10 with ResNet-18 across standard and plasticity-modulated optimizers.}
    (A) Test accuracy over training epochs.
    (B) Training loss over epochs.
    (C) Final test accuracy after 50 epochs.
    (D) Best test accuracy achieved during training.
    Standard optimizers (SGD, Adam, AdamW) achieve higher final performance under well-tuned learning rates, while NeuroPlastic becomes competitive at larger learning rates ($5\times10^{-2}$) but remains sensitive to optimization configuration. Despite this gap, NeuroPlastic exhibits stable convergence behavior across settings, suggesting that plasticity-modulated updates can provide a viable alternative optimization mechanism with further tuning or adaptive scheduling.
    }
    \label{fig:cifar10_gpu_validation}
\end{figure}

As shown in Figure~\ref{fig:cifar10_gpu_validation}, NeuroPlastic achieves competitive performance under appropriate learning rate settings, although it does not surpass well-tuned standard optimizers in this regime. The optimizer exhibits stable training dynamics across configurations but shows pronounced sensitivity to learning rate scaling. This behavior is consistent with the broader pattern observed throughout our experiments: the benefit of plasticity modulation depends on how well the optimization setting is matched to the modulation structure. In particular, NeuroPlastic's gains are tied to regimes where gradient information alone provides a weaker learning signal. At the same time, the fact that performance does not significantly degrade suggests that plasticity-modulated updates remain compatible with the underlying gradient optimization dynamics even outside its original tuning regime. This stability is useful in practice because it indicates that NeuroPlastic can be introduced into a training pipeline without destabilizing training across tasks. 

An additional transfer validation using the fixed MNIST-tuned configuration is provided in Appendix~\ref{appendix:cifar_transfer}, confirming that the performance gap relative to standard optimizers remains small and that training remains stable across configurations. To understand the sensitivity of these results to formulation choices, Appendix~\ref{appendix:design_space} includes a design space exploration evaluating a range of plasticity variants. Results show that performance varies substantially across formulations and that naive variants generally underperform standard optimizers, reinforcing the view that the specific modulation structure of NeuroPlastic matters and that plasticity is not a uniformly beneficial modification to gradient-based updates.

\section{Conclusions, Limitations, and Future Work}

NeuroPlastic optimizer provides modest but consistent improvements over a gradient-only ablation on lightweight benchmarks, with the clearest gains appearing on Fashion-MNIST and in low-data settings. Mechanistic diagnostics suggest that these improvements arise from modifying the effective update dynamics instead of substantially changing the raw gradient scale. These gains require no additional trainable parameters and no architectural modifications, making the approach a lightweight and compatible extension of standard gradient-based training. Taken together, our results suggest that biologically inspired plasticity signals can be instantiated as a a practical modulation layer on top of gradient updates, and point toward multi-signal modulation as a promising direction for extending gradient-based optimization in settings where learning signals are limited or noisy.

Some limitations remain in the current study. First, the experiments focus primarily on lightweight image classification benchmarks, limiting a more comprehensive characterization of NeuroPlastic's behavior. Second, the current plasticity formulation relies on a fixed modulation structure and manually chosen hyperparameters. Whether learned or adaptive weighting could improve performance, and under what conditions each signal contributes most, remains an open question. 

Future work could evaluate plasticity-modulated optimization on larger-scale architectures and datasets to fully explore NeuroPlastic's behavior and scalability. Additionally, future work could explore adaptive plasticity schedules, alternative plasticity signals, or mechanisms that automatically adjust modulation strength during training. Additional diagnostics also suggest that the plasticity coefficients can still exhibit boundary saturation in some settings, indicating room for improved constraint handling or adaptive range control.

\section*{Acknowledgment}
This work was supported by grants from the National Institutes of Health (R00 EY032181). We are grateful for Dr. Mu Qiao, Ms. Qiong Li and Ms. Yun Cao for insightful discussion on this work.

\section*{Conflict of Interest}
F. T. and D. J. are co-founders of Regenerative AI LLC. F. T. is a consultant of Netvirla Inc. The remaining authors declare no conflict of interest.

\bibliographystyle{plainnat}
\bibliography{references}

\appendix
\section*{Appendix}

\section{Optimizer Hyperparameters}

This appendix summarizes the default optimizer and stabilization hyperparameters used in the implementation of NeuroPlastic.

\subsection{Plasticity Configuration}

The default plasticity configuration is:
\begin{align}
w_a &= 0.4, &
w_g &= 0.4, &
w_m &= 0.2, \\
s_{\mathrm{plast}} &= 1.0, &
\alpha_{\min} &= 0.2, &
\alpha_{\max} &= 2.0.
\end{align}

The numerical stability constant is
\begin{equation}
\epsilon = 10^{-8}.
\end{equation}

By default, the optimizer uses:
\begin{itemize}
    \item rule-based plasticity mode,
    \item parameter-wise modulation,
    \item warmup disabled (\(\texttt{warmup\_epochs}=0\)),
    \item no learning-rate controller,
    \item no bounded residual branch,
    \item no orthogonal residual branch,
    \item no hybrid AdamW or SGD-momentum base.
\end{itemize}

\subsection{Trace and State Statistics}

The default activity-trace decay is
\begin{equation}
\beta_a = 0.95.
\end{equation}

The default momentum and variance decay coefficients are
\begin{equation}
\beta_1 = 0.9,
\qquad
\beta_2 = 0.99.
\end{equation}

\subsection{Homeostatic Stabilization}

The stabilization hyperparameters are:
\begin{align}
\tau &= 1.0, &
r_{\mathrm{target}} &= 0.02, &
\rho &= 0.01.
\end{align}

The gain is clipped to
\begin{equation}
\gamma_t^{\mathrm{stab}} \in [0.5,\,1.5].
\end{equation}

The implementation also supports layer-dependent target RMS scaling. The default values for early, middle, and late parameter groups are all set to
\begin{equation}
1.0.
\end{equation}

\subsection{Weight Decay and Learning Rate Control}

The optimizer supports both coupled and decoupled weight decay. The default value of weight decay is
\begin{equation}
0.0.
\end{equation}

The optional learning-rate controller is disabled by default. When enabled, its default parameters are:
\begin{align}
\alpha_{\mathrm{ctrl}} &= 0.1, &
\eta_{\min}^{\mathrm{ctrl}} &= 0.8, &
\eta_{\max}^{\mathrm{ctrl}} &= 1.2.
\end{align}

\section{Training Hyperparameters}

All experiments were conducted using identical model architectures and
training pipelines for both the full NeuroPlastic optimizer and the
gradient-only ablation within each experimental setting.

\textbf{Lightweight benchmarks (MNIST, Fashion-MNIST)}

\begin{itemize}
\item Optimizer learning rate: $\eta = 10^{-3}$
\item Batch size: 128
\item Training epochs: 10
\item Number of random seeds: 3
\item Plasticity coefficient bounds: $\alpha_{min}=0.9$, $\alpha_{max}=1.1$
\end{itemize}

\textbf{CIFAR-10 (ResNet-18, GPU validation)}

\begin{itemize}
\item Learning rates: $\{10^{-3}, 10^{-2}, 5\times10^{-2}\}$
\item Training epochs: 50
\item Batch size: 128
\item Optimizers compared: SGD, Adam, AdamW, NeuroPlastic
\item Number of random seeds: 3
\end{itemize}

\textbf{Datasets}

\begin{itemize}
\item MNIST
\item Fashion-MNIST
\item CIFAR-10
\end{itemize}

\section{Reproducibility}

To facilitate reproducibility, we provide the following experimental
protocol details:

\begin{itemize}

\item All models were trained using identical architectures and training
pipelines across optimizers within each experimental setting.

\item Random seeds were fixed for dataset splitting, weight initialization,
and training order.

\item Each experiment was repeated across three independent seeds and
results are reported as mean performance with standard deviation.

\item The gradient-only ablation is implemented by removing the activity and memory contributions while retaining gradient-based modulation:
\[
\alpha_t^{\mathrm{grad}} = \mathrm{clip}\!\left(g_t^{(s)}, \alpha_{\min}, \alpha_{\max}\right).
\]

\item NeuroPlastic is implemented in PyTorch as a drop-in optimizer without
modifying model architectures or introducing additional trainable parameters.

\item The plasticity coefficient $\alpha_t$ is computed per-parameter and
normalized as defined in the main formulation.

\item Stabilization is applied to all updates to prevent instability,
particularly at larger learning rates.

\item Learning rate sensitivity is explicitly evaluated, and multiple
learning rates are reported to reflect different optimization regimes.

\end{itemize}

\section{Compute Resources}
All experiments were conducted using CPU or NVIDIA GPU in a standard workstation environment using PyTorch. Depending on the dataset and model configuration, runtime ranges from 1 to 24 hours. The overall compute used during development is comparable in scale to the experiments reported.

\section{Additional Transfer Validation on CIFAR-10}
\label{appendix:cifar_transfer}
\begin{figure}[t]
\centering
\includegraphics[width=\linewidth]{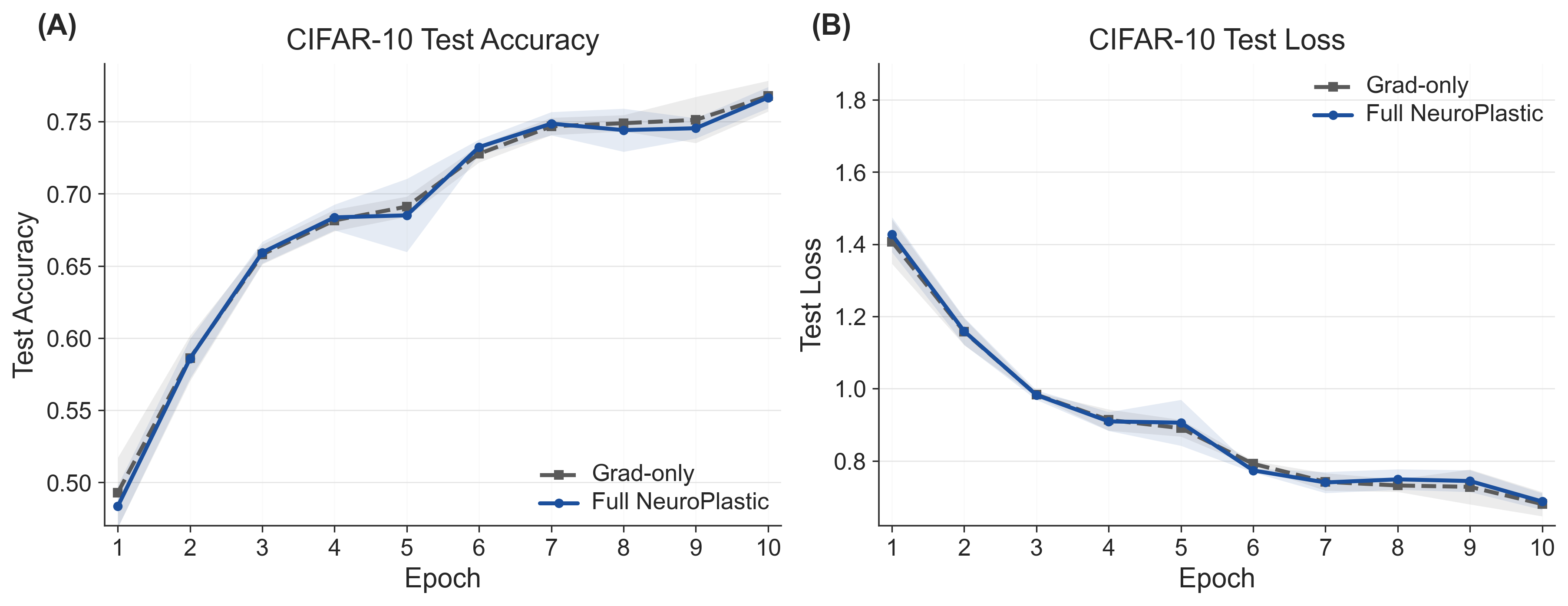}
\caption{
\textbf{Transfer validation on CIFAR-10.}
(A) Test accuracy across training epochs when applying the optimizer configuration tuned on MNIST.
The gradient-only baseline achieves slightly higher final accuracy.
(B) Test loss across epochs.
Both optimizers exhibit stable convergence behavior, indicating that the NeuroPlastic mechanism does not destabilize training when transferred across datasets.
Shaded regions denote the standard deviation across independent runs.
}
\label{fig:appendix_cifar_transfer}
\end{figure}

We further evaluate the transfer behavior of the optimizer on CIFAR-10 using the fixed hyperparameter configuration obtained from MNIST tuning. This experiment serves as a supplementary validation to assess whether the plasticity-modulated optimizer generalizes to a more complex dataset without additional retuning.

Under this setting, the gradient-only ablation achieves slightly higher final accuracy than the full NeuroPlastic optimizer, with an average difference of approximately 0.001 (Appendix Figure~\ref{fig:appendix_cifar_transfer}). While the full optimizer occasionally attains comparable or slightly higher intermediate accuracy during training, this advantage is not consistently maintained at convergence.

Overall, these results indicate that the MNIST-derived configuration does not directly translate into improved performance on CIFAR-10. Nevertheless, the performance gap remains small, suggesting that the plasticity-modulated update does not introduce instability when applied outside its original tuning regime. Additional transfer results are shown in Appendix Figure~\ref{fig:appendix_cifar_transfer}.

\section{Analysis of Plasticity-Modulated Optimization Behavior}
\label{appendix:mechanistic_analysis}
\begin{figure}[t]
\centering
\includegraphics[width=\linewidth]{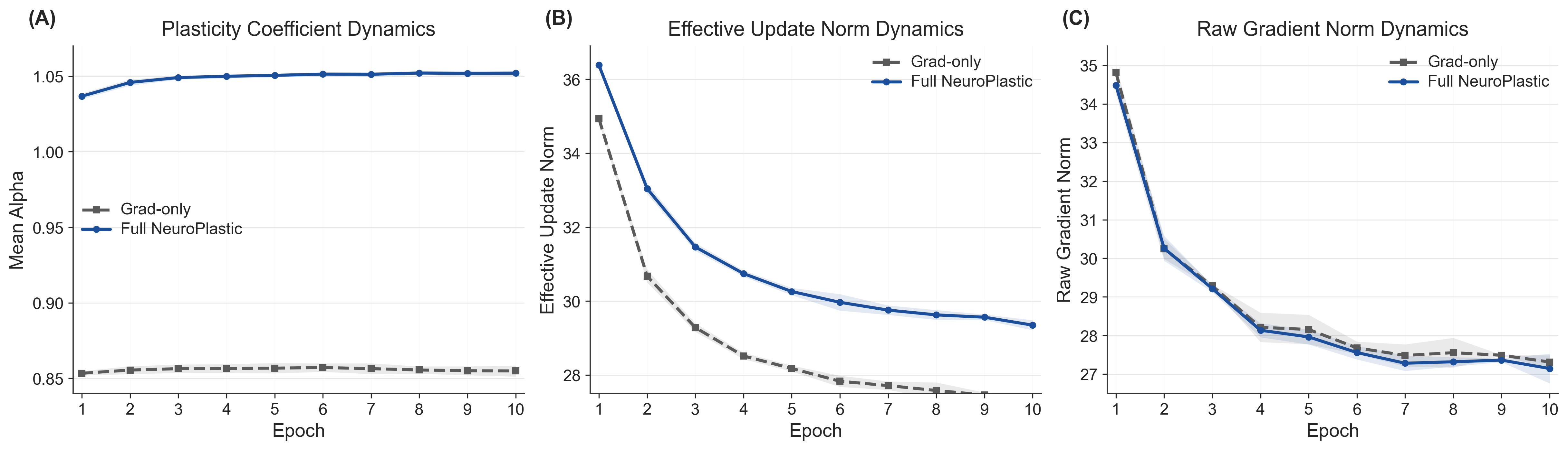}
\caption{
  \textbf{Mechanistic optimizer diagnostics on Fashion-MNIST.}
  (A) Mean plasticity coefficient dynamics.
  Full NeuroPlastic maintains a higher and more stable effective plasticity level than the gradient-only condition.
  (B) Effective update norm dynamics.
  Full NeuroPlastic exhibits a distinct update trajectory, with slightly larger early effective updates and a smoother
  decay over training.
  (C) Raw gradient norm dynamics.
  The raw gradient scale remains similar between the two conditions, indicating that the difference in optimization
  behavior is not primarily driven by larger gradients, but by how updates are transformed and applied.
  Solid blue lines denote Full NeuroPlastic; dashed gray lines denote gradient-only.
  Shaded bands indicate $\pm 1$ standard deviation over three random seeds.
}
\label{fig:mechanistic_diagnostics}
\end{figure}

To characterize the effect of plasticity modulation, we analyze optimizer state dynamics during training on Fashion-MNIST (Appendix Figure~\ref{fig:mechanistic_diagnostics}). NeuroPlastic maintains a higher and more stable plasticity coefficient than the gradient-only ablation, indicating sustained modulation throughout training. The effective update norm follows a distinct trajectory under NeuroPlastic, with slightly larger updates during early training and a smoother decay in later stages. This suggests that improvements arise from modifying update dynamics rather than gradient scale.

We further observe that removing normalization or stabilization can lead to unstable training dynamics, particularly at larger learning rates, where the plasticity coefficient may grow excessively and produce exploding updates. These observations highlight the importance of bounded modulation and normalization for maintaining stable optimization behavior.

Taken together, these results support the view that NeuroPlastic functions as a secondary modulation layer on top of gradient-based optimization, altering effective update dynamics without substantially changing the underlying gradient scale.

\section{Design Space Exploration of Plasticity-Modulated Optimization}
\label{appendix:design_space}
To further understand how different formulations of plasticity-modulated updates affect optimization behavior, we evaluate a range of variants derived from the proposed framework. These include residual-style updates, hybrid modulation schemes, and an interpretation of plasticity as a learning-rate controller.

As shown in Appendix Fig.~\ref{fig:design_space_summary}, performance varies substantially across formulations. Naive variants such as residual or hybrid modulation generally underperform standard optimizers, indicating that directly injecting plasticity signals into the update may not consistently improve optimization.

\begin{figure}[ht]
\centering
\includegraphics[width=\linewidth]{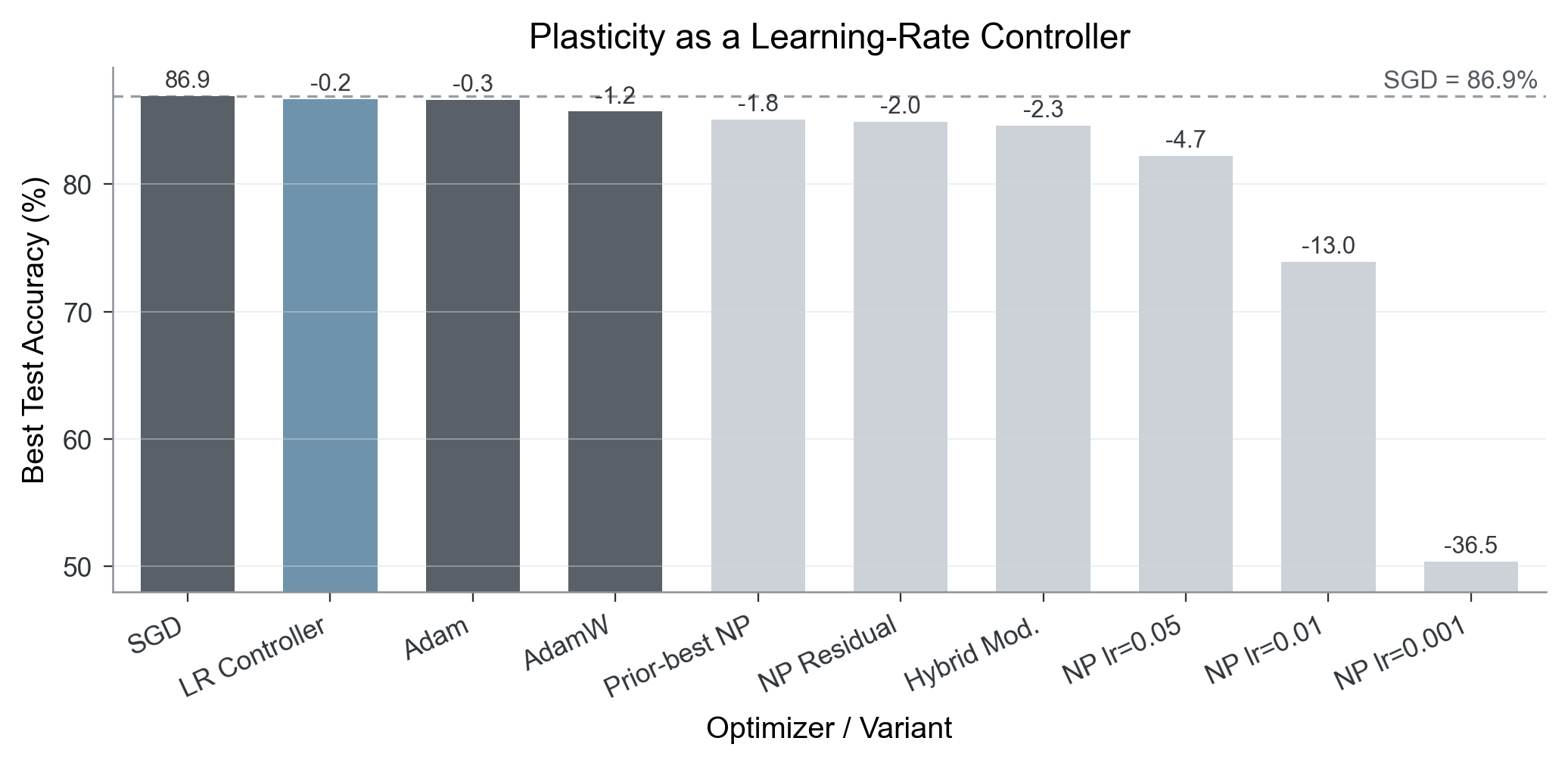}
\caption{
Design space exploration of plasticity-modulated optimization variants on CIFAR-10 (ResNet-18). Each bar shows the best test accuracy achieved by a given optimizer or formulation. Differences relative to the SGD baseline are annotated above the bars. While several naive variants underperform, a controller-style formulation of plasticity achieves performance close to SGD, suggesting that the effectiveness of plasticity modulation depends strongly on how it is incorporated into the update rule.
}
\label{fig:design_space_summary}
\end{figure}

In contrast, a controller-style interpretation, in which plasticity modulates the effective learning rate rather than acting as an additive update component, achieves performance close to well-tuned SGD. This suggests that the effectiveness of plasticity-modulated updates depends critically on how the modulation signal is integrated into the optimization process.

These observations highlight that plasticity is better understood as a modulation mechanism whose impact is highly formulation-dependent, rather than as a standalone improvement over existing optimizers. Identifying formulations that align with stable and effective optimization dynamics remains an important direction for future work.

%\clearpage
%\input{checklist.tex}

\end{document}